\documentclass[10pt,twocolumn,letterpaper]{article}

\usepackage{iccv}

\usepackage{times}
\usepackage{epsfig}
\usepackage{graphicx}
\usepackage{amsmath}
\usepackage{amssymb}
\usepackage{graphicx}
\usepackage{comment}
\usepackage{amsmath,amssymb} 
\usepackage{color}
\usepackage{adjustbox}
\usepackage{caption}
\usepackage{multirow}
\usepackage{tabu}           
\usepackage{multirow}
\usepackage{booktabs}
\usepackage{makecell}
\usepackage{pifont}

\usepackage{color}
\usepackage{colortbl}
\usepackage{color}
\usepackage{pifont}

\makeatletter
\@namedef{ver@everyshi.sty}{}
\makeatother
\usepackage{tikz}
\usetikzlibrary{arrows,automata,positioning}
\definecolor{Gray}{gray}{0.95}
\definecolor{Cyan}{rgb}{0.88,1,1}

\usepackage{algorithm}
\usepackage{algorithmicx}
\usepackage{algpseudocode}

\usepackage[caption=false]{subfig}
\usepackage[pagebackref=true,breaklinks=true,colorlinks,bookmarks=false]{hyperref}

\iccvfinalcopy 

\usepackage[caption=false]{subfig}
\usepackage[pagebackref=true,breaklinks=true,colorlinks,bookmarks=false]{hyperref}

\usepackage{listings}

\definecolor{codegreen}{rgb}{0,0.6,0}
\definecolor{codegray}{rgb}{0.5,0.5,0.5}
\definecolor{codepurple}{rgb}{0.58,0,0.82}
\definecolor{backcolour}{rgb}{0.95,0.95,0.92}

\lstdefinestyle{mystyle}{
    backgroundcolor=\color{backcolour},   
    commentstyle=\color{codegreen},
    keywordstyle=\color{magenta},
    numberstyle=\tiny\color{codegray},
    stringstyle=\color{codepurple},
    basicstyle=\ttfamily\footnotesize,
    breakatwhitespace=false,         
    breaklines=true,                 
    captionpos=b,                    
    keepspaces=true,                 
    numbers=left,                    
    numbersep=5pt,                  
    showspaces=false,                
    showstringspaces=false,
    showtabs=false,                  
    tabsize=2
}

\newcolumntype{x}[1]{>{\centering\arraybackslash}p{#1pt}}
\newlength\savewidth\newcommand\shline{\noalign{\global\savewidth\arrayrulewidth
  \global\arrayrulewidth 1pt}\hline\noalign{\global\arrayrulewidth\savewidth}}

\newcommand{\tablestyle}[2]{\setlength{\tabcolsep}{#1}\renewcommand{\arraystretch}{#2}\centering\footnotesize}

\def\iccvPaperID{7576} 

\ificcvfinal\fi

\begin{document}

\title{RandomRooms: Unsupervised Pre-training from Synthetic Shapes and Randomized Layouts for 3D Object Detection}

\author{Yongming Rao\textsuperscript{1}\thanks{Equal contribution. ~\textsuperscript{\dag}Corresponding author.}, ~Benlin Liu\textsuperscript{2,3}\footnotemark[1], ~Yi Wei\textsuperscript{1}, ~Jiwen Lu\textsuperscript{1}\textsuperscript{\dag}, ~Cho-Jui Hsieh\textsuperscript{2}, ~Jie Zhou\textsuperscript{1}\\
\textsuperscript{1}Tsinghua University,~\textsuperscript{2}UCLA,~\textsuperscript{3}University of Washington \\
{\tt\small  raoyongming95@gmail.com; liubl@cs.washington.edu; wziyi20@mails.tsinghua.edu.cn; } \\
{\tt\small  chohsieh@cs.ucla.edu; \{lujiwen, jzhou\}@tsinghua.edu.cn} \\
}

\maketitle

\begin{abstract}
3D point cloud understanding has made great progress in recent years. However, one major bottleneck is the scarcity of annotated real datasets, especially compared to 2D object detection tasks, since a large amount of labor is involved in annotating the real scans of a scene. A promising solution to this problem is to make better use of the synthetic dataset, which consists of CAD object models, to boost the learning on real datasets. This can be achieved by the pre-training and fine-tuning procedure. However, recent work on 3D pre-training exhibits failure when transfer features learned on synthetic objects to other real-world applications. In this work, we put forward a new method called RandomRooms to accomplish this objective. In particular, we propose to generate random layouts of a scene by making use of the objects in the synthetic CAD dataset and learn the 3D scene representation by applying object-level contrastive learning on two random scenes generated from the same set of synthetic objects. The model pre-trained in this way can serve as a better initialization when later fine-tuning on the 3D object detection task. Empirically, we show consistent improvement in downstream 3D detection tasks on several base models, especially when less training data are used, which strongly demonstrates the effectiveness and generalization of our method. Benefiting from the rich semantic knowledge and diverse objects from synthetic data, our method establishes the new state-of-the-art on widely-used 3D detection benchmarks ScanNetV2 and SUN RGB-D. We expect our attempt to provide a new perspective for bridging object and scene-level 3D understanding.  

\end{abstract}

\section{Introduction}

\begin{figure}
  \centering
  \includegraphics[width=\linewidth]{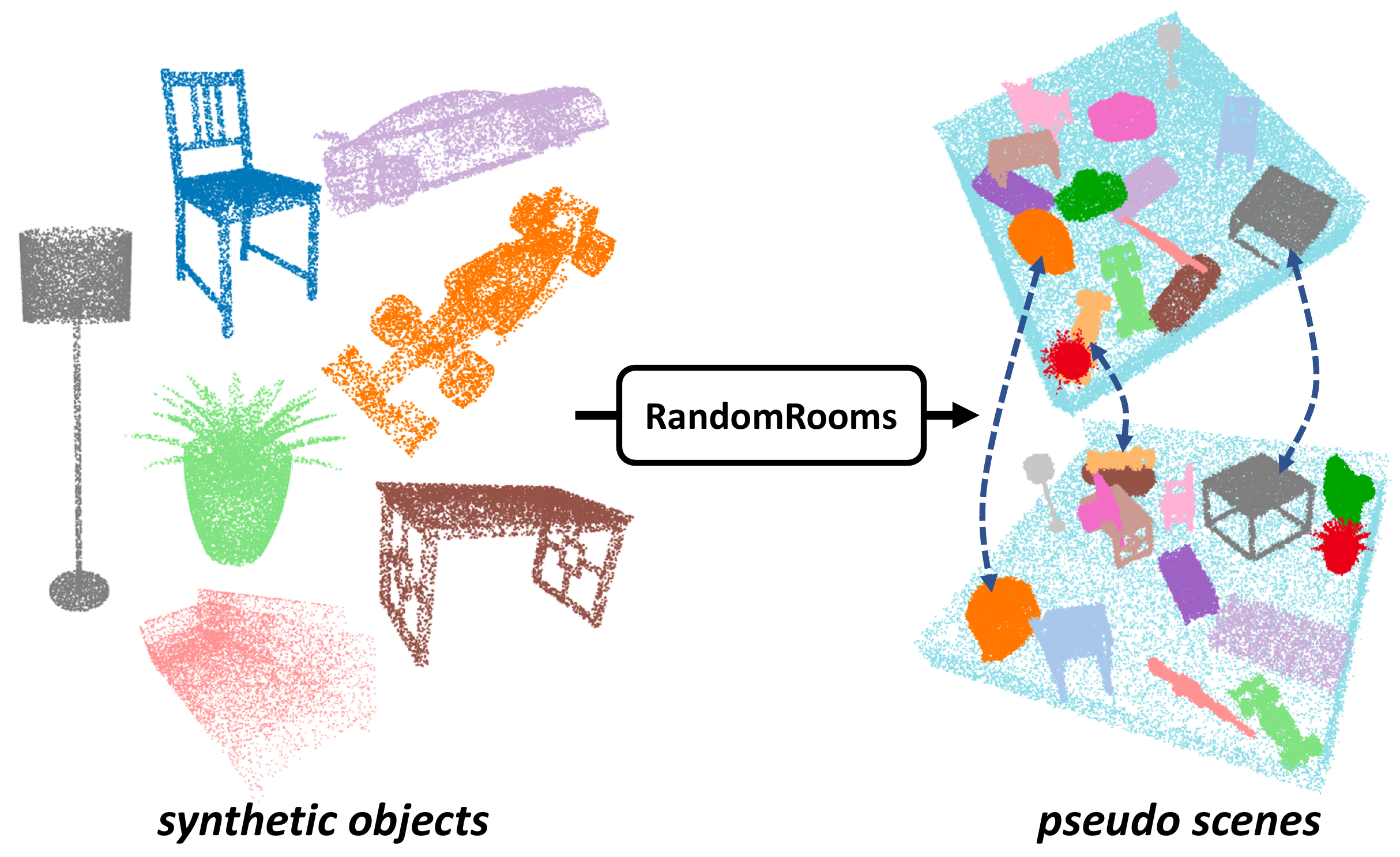}
  \caption{The main idea of \textit{RandomRooms}. To generate two different layouts, we randomly place the same set of objects sampled from synthetic datasets in rectangular rooms. With the proposed object-level contrastive learning, models pre-trained on these pseudo scenes can serve as a better initialization for downstream 3D object detection task.}
  \label{fig:intro}
  \vspace{-10pt}
\end{figure}

Recent years have witnessed great progress in 3D deep learning, especially on 3D point clouds. With the emergence of powerful models, we are now able to make significant breakthroughs on many point cloud tasks, ranging from object-level understanding ones~\cite{klokov2017escape,dgcnn,so-net-li2018so,li2018pointcnn} to scene-level understanding ones, such as 3D object detection~\cite{shi2019pointrcnn,yang20203dssd,li2019stereo,shi2019pv} and 3D semantic segmentation~\cite{kundu2020virtual,zhang12356deep,hu2020jsenet,choy20194d,hou20193d}. These scene-level tasks are considered to be more complicated and more important as they often require higher level understanding compared to object level tasks like shape classification. One of the most important tasks for 3D point cloud scene understanding is the 3D object detection, which aims at localizing the objects of interest in the point cloud of the scene and telling the category they belong to. However, one major bottleneck that hinders the researchers from moving forward is the lack of large-scale real datasets, considering the difficulty in collecting and labeling high-quality 3D scene data. Compared to 2D object detection task where we have large annotated real datasets COCO~\cite{lin2014microsoft}, the real datasets here we use for 3D object detection task are much smaller in scales, and generating a synthesized scene dataset also involves a heavy workload in modeling and rendering.

A preferred solution is to utilize synthetic CAD object models to help the learning of 3D object detector since it is much easier to access such type of data. Considering we have no annotation of bounding box for synthetic CAD data, this idea can be achieved in a similar way as the unsupervised pre-training for 2D vision tasks where we first pre-train on a large-scale dataset in an unsupervised manner and then fine-tune on a smaller annotated dataset. Yet, most previous works focus on the pre-training for single object level tasks~\cite{L2G-liu2019l2g,yang2018foldingnet,deng2018ppf,gadelha2018multiresolution,pointglr}, such as reconstruction, shape classification or part segmentation, or on some low-level tasks like registration~\cite{deng2018ppf,zeng20173dmatch,elbaz20173d}. A recent work~\cite{xie2020pointcontrast}, namely PointContrast, first explores the possibility of pre-training in the context of 3D representation learning for higher level scene understanding tasks, i.e. 3D detection and segmentation. Nevertheless, they conduct the pre-training on the real scene dataset and provide a failure case when pre-training the backbone model on ShapeNet~\cite{chang2015shapenet}, which consists of synthetic CAD object models. They attribute this unsuccessful attempt to two reasons, that is, the domain gap between real and synthetic data as well as the insufficiency of capturing point-level representation by directly training on single objects. Despite these difficulties, it is still desirable to make the ShapeNet play the role of ImageNet in 2D vision since it is easy to obtain a large number of synthetic CAD models. 

In this work, we put forward a new framework to show the possibility of using a synthetic CAD model dataset, i.e. ShapeNet, for the 3D pre-training before fine-tuning on downstream 3D object detection task. To this end, we propose a method named RandomRoom. In particular,  we propose to generate two different layouts using one set of objects which are randomly sampled out of the ShapeNet dataset. Having these two scenes that are made up of the same set of objects, we can then perform the contrastive learning at the object level to learn the 3D scene representation. 

Different from PointContrast~\cite{xie2020pointcontrast} where the contrastive learning is performed at the point level, our approach has two advantages. One is to remove the requirement of point correspondence between two views, which is indispensable in PointContrast framework given that it is necessary to exploit such information to obtain positive and negative pairs for the contrastive learning. This requirement limits  the applications of PointContrast, since the CAD model datasets like ShapeNet and many other real-world datasets like SUN RGB-D~\cite{song2015sun} cannot provide such information. 
The other advantage is that our method can support more diverse backbone models. Most state-of-the-art models~\cite{puy2020flot,qi2019deep,shi2019pv} on tasks like 3D object detection apply PointNet++~\cite{qi2017pointnet++} style models as their backbone, and replacing it with Sparse Res-UNet may lead to the drop of accuracy, according to the PointContrast. However, PointContrast cannot well support the pre-training of PointNet++ style model as the UNet-like models, since the point correspondence may be missing after each abstraction level in PointNet++. With the proposed RandomRoom, we are enabled to perform contrastive learning at the level of objects 
and thus better support the pre-training of PointNet++ like models as we no longer need to  keep the point correspondence for contrastive learning like PointContrast.

Our method is straightforward yet effective. We conduct the experiments on the 3D object detection task where only the geometric information is available for input as the models in CAD datasets do not carry color information. The results of empirical study strongly demonstrate the effectiveness of our method. In particular, we achieve the state-of-the-art of 3D object detection on two widely-used benchmarks, ScanNetV2 and SUN-RGBD. Furthermore, our method can achieve even more improvements when much less training samples are used, demonstrating that our model can  learn a better initialization for 3D object detection.

\section{Related Work}

\begin{figure*}
  \centering
  \includegraphics[width=0.85\linewidth]{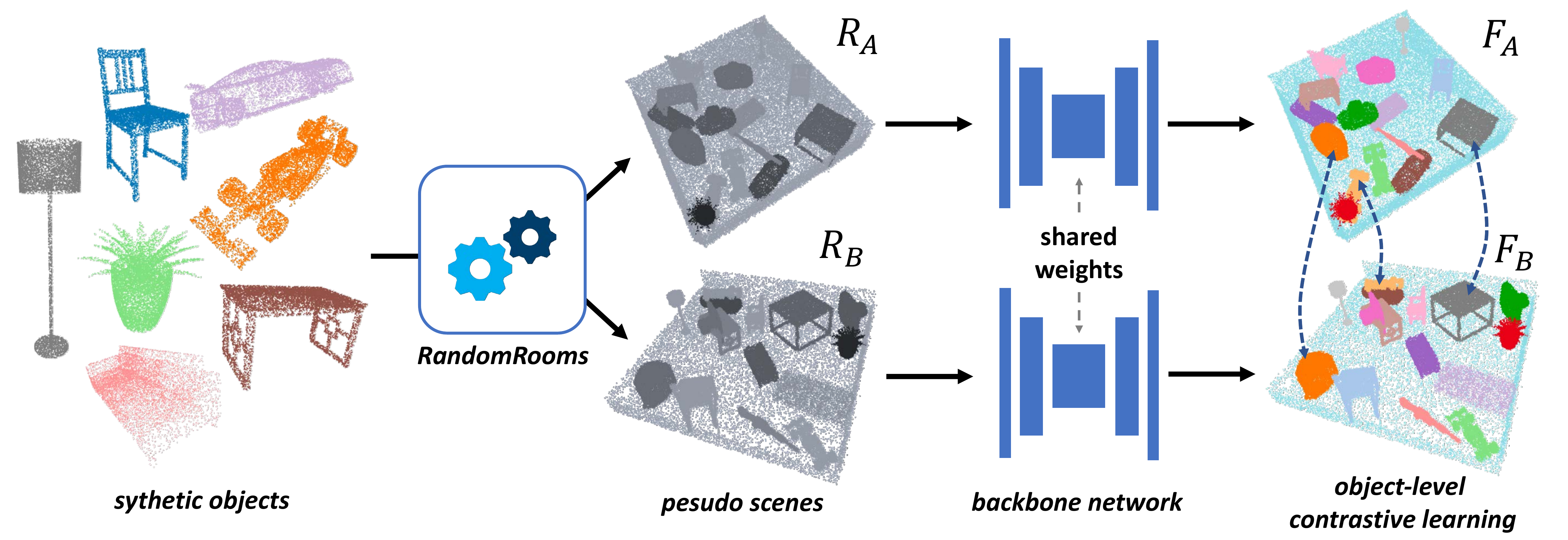}
  \caption{The overview of our framework. Given the objects randomly sampled from synthetic datasets, pairs of pesudo scenes are constructed following object augmentation, layout generation and scene augmentation. We pretrain the model with shared weights on two corresponding random rooms. An object-level contrastive  learning (OCL) method is proposed to help the network learn discriminative representation. }
  \label{fig:method}
  \vspace{-10pt}
\end{figure*}

\noindent \textbf{3D Deep Learning. }
3D deep learning~\cite{jiang2020pointgroup,pointglr,shi2019pointrcnn,yang20203dssd,li2019stereo,shi2019pv,hou20193d,wang2018sgpn,wei2019conditional} has attracted much attention in recent years, especially on 3D point cloud analysis~\cite{qi2017pointnet,qi2017pointnet++,klokov2017escape,dgcnn,so-net-li2018so,li2018pointcnn}. As the pioneer work, PointNet \cite{qi2017pointnet} introduces deep learning to 3D point cloud analysis . With the max pooling layer, it is able to directly operate on unordered set. As a follow up, PointNet++ \cite{qi2017pointnet++} employs PointNet as a basic module to hierachically extract features. Different from~\cite{qi2017pointnet,qi2017pointnet++}. Many other variants of PointNet++ are also devised to further improve feature capacity~\cite{klokov2017escape,thomas2019kpconv}. Thanks to these architectures, significant processes have been made in many 3D applications \cite{klokov2017escape,dgcnn,so-net-li2018so,li2018pointcnn,shi2019pv,qi2019deep,hou20193d,wang2018sgpn}. As the data-driven methods, these works either use object-level synthetic training data or leverage point clouds from real scenes. Exploring the great power of both synthetic and real-world datasets, our method bridges the gaps between object and scene level 3D understanding. 

\vspace{5pt} \noindent \textbf{3D Object Detection. }
Due to the broad real-world applications, more and more works \cite{shi2019pointrcnn,yang20203dssd,li2019stereo,shi2019pv,hou20193d,wang2018sgpn,xie2020pointcontrast} focus on 3D scene understanding. As a fundamental 3D task, 3D object detection focuses on the problem of detecting objects' tight bounding boxes in 3D space. F-PointNet \cite{qi2018frustum} predicts 3D bounding boxes from the points in frustums and achieves efficiency as well as high recall for small objects. It can also handle strong occlusion or cases with very sparse points. Inspired by Hough voting process, VoteNet \cite{qi2019deep} leverages voting mechanism to capture scene context around objects centers. Based on VoteNet, H3DNet \cite{zhang2020h3dnet} predicts different modalities of geometric primitives and aggregate them to generate final 3D bounding boxes. Benefiting from hybrid features, H3DNet achieves state-of-the-art performance.  However, these 3D scene understanding methods mainly make use of the real data from 3D sensors. On the contrary, our method aims at bringing the semantic knowledge in synthetic datasets to high-level 3D understanding tasks. 

\vspace{5pt} \noindent \textbf{Model Pre-training. }
Pre-training has been the common practice for many machine learning tasks, ranging from vision~\cite{xie2020pointcontrast,simclr,moco,mocov2,infomin,girshick2014rich} to NLP tasks~\cite{peters2018deep,radford2018improving,howard2018universal,devlin2019bert}. In the context of 2D vision, the pre-training is often conducted on ImageNet~\cite{deng2009imagenet} with full supervision, and we can then fine-tune the pre-trained backbone model on downstream tasks like detection~\cite{girshick2014rich,ren2015faster,girshick2015fast}. More recently, unsupervised pre-training on ImageNet~\cite{simclr,moco,mocov2} has been been showed to be effective. Compared to 2D vision, less exploration has been made on 3D vision tasks. Previously, most methods on 3D pre-training either focus on the tasks at single object level, like classification, reconstruction and part segmentation~\cite{yang2018foldingnet,gadelha2018multiresolution,pointglr,hassani2019unsupervised}, or on some low-level 3D tasks like registration~\cite{deng2018ppf,zeng20173dmatch,elbaz20173d}. Pre-training for higher level 3D scene understanding tasks like detection and segmentation has not been studied only until a recent work~\cite{xie2020pointcontrast}, which exploits the point correspondence to learn the representation in an unsupervised manner. Compared to theirs, our method can pre-train on synthetic CAD datasets like ShapeNet and support more types of backbone model.


\section{RandomRooms}

In this section, we describe the details of the proposed RandomRooms method. We first briefly review existing contrastive representation learning methods and illustrate the intuition of our method in Section~\ref{sec:CLR}. Then, we describe how to use synthetic objects to construct random rooms in~\ref{sec:construction}. In Section~\ref{sec:learning}, we show our pretrain task for learning scene level representation from the pseudo scenes. The overview of our framework is presented in Figure~\ref{fig:method}.

\subsection{Overview of Contrastive Learning}\label{sec:CLR}
We begin by reviewing the existing contrastive representation learning methods for 2D and 3D understanding to illustrate the motivation of our method. 

Contrastive learning is at the core of several recent methods on unsupervised learning, which exhibits promising performance on both 2D~\cite{instance-dis-wu2018unsupervised,cpc-henaff2019data,tian2019contrastive,moco,mocov2,simclr,byol,infomin} and 3D~\cite{xie2020pointcontrast,pointglr} tasks and shows impressive generalization ability as a new type of pre-training method for various downstream tasks. The key ingredient of contrastive learning is constructing positive and negative pairs to learn discriminative representation, which inherits the idea of conventional contrastive learning in metric learning literature~\cite{hadsell2006dimensionality}. Given an input $x$ and its positive pair $x_+$ and a set of negative examples $\{x_i\}$, a commonly used training objective for contrastive representation learning is based on InfoNCE~\cite{cpc-henaff2019data,tian2019contrastive}:
\begin{equation}
    \mathcal{L}_\text{contrastive} = -\log \frac{\exp(\varphi(x) \cdot \varphi(x_+)/\tau)}{\sum_i \exp(\varphi(x) \cdot \varphi(x_i)/\tau)},
\end{equation}
where $\varphi$ is the encoder network that maps the input to a feature vector and $\tau$ is a temperature hyper-parameter  following~\cite{instance-dis-wu2018unsupervised,moco,simclr}.  Intuitively, the contrastive learning methods supervise models by encouraging the features of the different \textit{views} of the same sample to be close to each other and distinguishable from other samples~\cite{n-pair-sohn2016improved,schroff2015facenet}. Hence the quality of positive pairs and negative examples is a critical factor to learn the encoder. 

Since category annotations are not available in the unsupervised learning scenario, a common practice~\cite{examplar,instance-dis-wu2018unsupervised, moco} is using different augmentations of an input as the positive pairs and treating all other samples as negative examples.  Although this design has proven to be effective in image representation learning, we argue there is a better solution to construct positive pairs for 3D understanding. One fundamental difference between 2D and 3D data is that the spatial structures of pixels do not reflect the actual geometric structures of the objects, but the spatial structures in 3D data always faithfully illustrate the layouts in the real world. This property suggests that it may be easier to manipulate or \textit{augment} 3D data compared to 2D images. Inspired by the rendering techniques in computer graphics, we propose to generate positive pairs of 3D scenes by randomly manipulating the layouts of 3D objects in a scene. Since we only need 3D objects instead of the whole scene in this process, our method makes it possible to use 3D object models to promote scene level representation learning. 

It is worth noting that a recent work, namely PointContrast~\cite{xie2020pointcontrast}, explores 3D contrastive  representation learning by using 3D point clouds from different views as the positive pair, where a point level contrastive loss is designed. This method is based on the multi-view point cloud sequences provided in ScanNetV2~\cite{dai2017scannet}. Instead, our method focuses on leveraging object level 3D data, which are easier to collect and have more diverse categories. 

\subsection{Random Rooms from Synthetic Objects}\label{sec:construction}

\begin{figure}
  \centering
  \includegraphics[width=\linewidth]{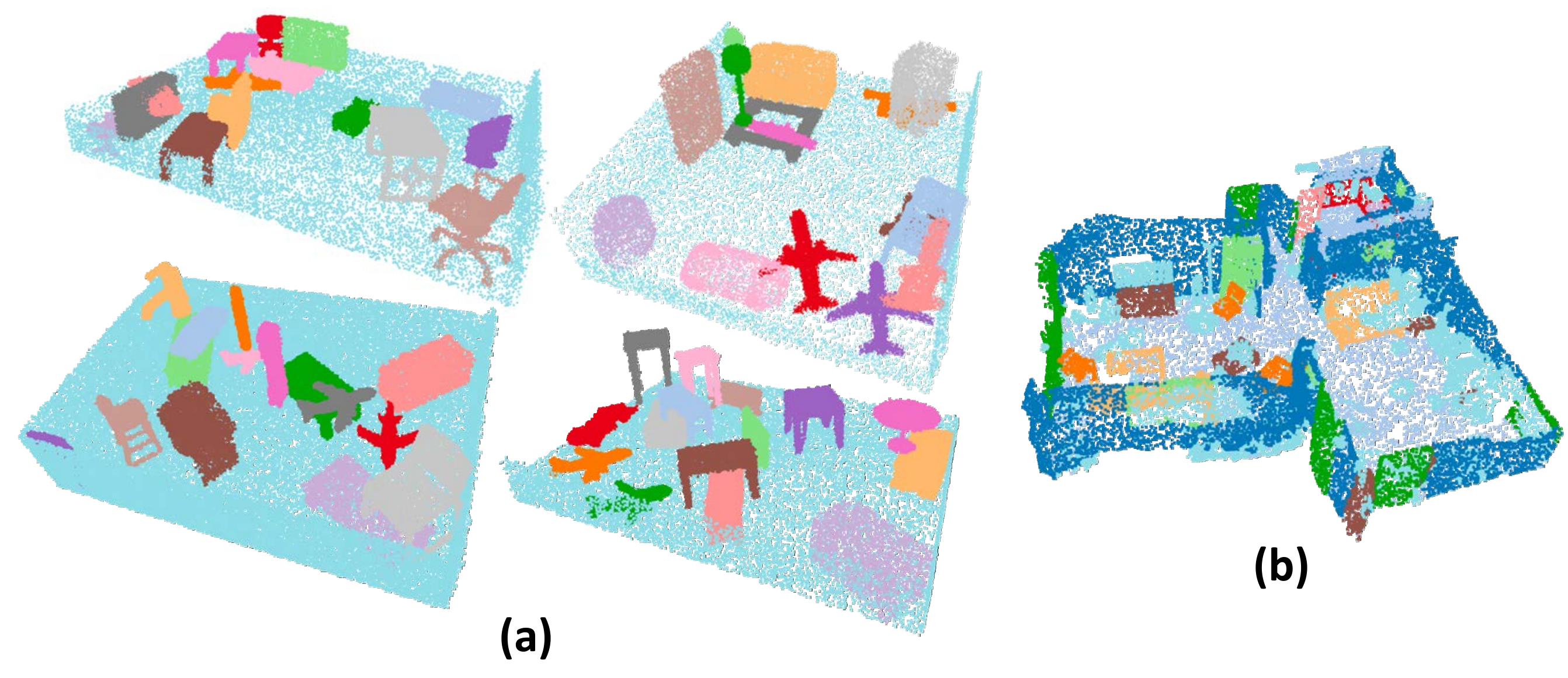}
  \caption{Some randomly selected examples of random rooms (a) and scene from ScanNetV2 (b). }
  \label{fig:example}
  \vspace{-10pt}
\end{figure}

Compared to ScanNetV2~\cite{dai2017scannet}, which contains $\sim$15k objects from 17 categories, synthetic shape datasets like ShapeNet~\cite{wu2015shapenet} provide a more plentiful source for 3D understanding. For example, ShapeNetCore~\cite{wu2015shapenet} contains $\sim$52k objects from 55 categories). Therefore, the primary goal of this paper is to study how to use synthetic CAD models collected by ShapeNet to improve downstream tasks like 3D detection and segmentation on real-world datasets.

Previous work~\cite{xie2020pointcontrast} shows that directly pre-training on ShapeNet will not yield performance improvement on downstream detection and segmentation task. We suspect the main reason is the domain gap between the single object classification task on ShapeNet and the multiple objects localization task on real-world datasets. In order to bridge the gap,  we propose to generate pseudo scenes (we name them as \textbf{\textit{random rooms}}) from synthetic objects to construct the training data that are helpful for scene level understanding. 

Given a set of randomly sampled objects, we generate a random room following the three steps:
\begin{itemize}
    \item \textbf{Object Augmentation:} We first resize the object to a random size in [0.5m, 2.0m] to ensure the objects have similar sizes as the objects in ScanNetV2. Then, we apply commonly used object point cloud augmentation techniques~\cite{qi2017pointnet,qi2017pointnet++,rscnn-liu2019relation} including rotation, point dropping, jittering. 
    \item \textbf{Layout Generation:} For the ease of implementation, we place objects in a rectangular room. The size of the room is adaptively adjusted according to the overall area of the augmented objects. The layout is generated based on two simple principles: 1) non-overlapping: any two objects should not occupy the same space in the room; 2) gravity: objects should not float in the air, and larger objects should not be placed over the smaller ones. In turn, we place objects in the descending order of the area.  Inspired by \textit{Tetris}~\footnote{https://en.wikipedia.org/wiki/Tetris}, for each object, we first randomly choose a position in the X-Y plane that satisfies the above principles, then determine the location (the Z value) based on the current maximum height of the position. The object will not be placed in a position if the current maximum height of the position exceeds 2m. 
    \item \textbf{Scene Augmentation:} Lastly, we apply data augmentation like rotation along the Z axis, point dropping, jittering to the whole scene. To make the generated scenes more similar to the real scenes, we also add the floor and walls as confounders.
\end{itemize}

Some examples of the random rooms are illustrated in Figure~\ref{fig:example}. 

\subsection{Representation Learning from Random Rooms}\label{sec:learning}

To utilize the generated random rooms, we devise an object-level contrastive learning (OCL) method, which learns discriminative representation without category annotations. 

Given $n$ randomly sampled objects $\{x_1,x_2,...,x_n\}$, we first generate two random rooms $R_A = \{x_1^A, x_2^A, ..., x_n^A \}$ and $R_B = \{x_1^B, x_2^B, ..., x_n^B \}$ by conducting the above-mentioned steps individually. Then, we employ the point cloud encoder-decoder network $\mathcal{M}$ (\eg PointNet++~\cite{qi2017pointnet++} with feature propagation layers) to extract per-point features of the two scenes $F_A = \mathcal{M}(R_A)$ and  $F_B = \mathcal{M}(R_B)$. Since the random room is constructed by several individual objects, the instance labels can be naturally defined. The goal of object-level contrastive learning is to exploit the instance labels as a source of free and plentiful supervisory signals for training a rich representation for point cloud understanding. To obtain the feature of each object, we apply the average pooling operation $\mathcal{A}$ on per-point features belonging to this object:
\begin{equation}
    \{h_1^A, h_2^A, ..., h_n^A\}= \mathcal{A}(F_A), \quad  \{h_1^B, h_2^B, ..., h_n^B\}= \mathcal{A}(F_B). \nonumber
\end{equation}
Similar to the common practice in contrastive learning~\cite{mocov2,simclr}, the object features then are projected onto a unit hypersphere using a multi-layer perceptron network (MLP) followed by L2 normalization. The object-level contrastive learning objective can be written as:
\begin{equation}
\begin{split}
    \mathcal{L}_\text{OCL} = &- \frac{1}{n} \sum_{i=1}^n  \log \frac{\exp(f_i^A \cdot f_i^B/\tau)}{\sum_{f\in \mathcal{F}} \exp(f_i^A  \cdot f/\tau)} \\
    &- \frac{1}{n} \sum_{i=1}^n  \log \frac{\exp(f_i^B \cdot f_i^A/\tau)}{\sum_{f\in \mathcal{F}} \exp(f_i^B  \cdot f/\tau)} ,
\end{split}
\end{equation}
where $f_i^A = \phi(h_i^A)$ and $f_i^B = \phi(h_i^B)$ are the projected features of the $i$-th object in $R_A$ and $R_B$ respectively, $\phi$ is the projection head, and $\mathcal{F}$ is the set of all projected features in the batch. Note that compared to point-level contrastive learning task in PointContrast~\cite{xie2020pointcontrast}, our method further utilizes the instance-level knowledge thanks to the generation mechanism of RandomRooms. We argue that object-level contrastive learning introduces more semantic knowledge and can be more helpful for downstream localization tasks (Some empirical evidence can be found in Table~\ref{tb:method}).

\section{Experiments}

One primary goal of representation learning is to learn representation that can transfer to downstream tasks. To apply our RandomRooms method to scene level understanding task like 3D object detection, we adopt the \textit{unsupervised pre-training} + \textit{supervised fine-tuning} pipeline~\cite{moco,xie2020pointcontrast}. Specifically, we first pre-train the backbone model on ShapeNet using our method, then we use the pre-trained weights as the initialization and further fine-tune the model on the downstream 3D object detection task. 

\subsection{Pre-training Setups}
We perform the pre-training on ShapeNet~\cite{chang2015shapenet}, a dataset composed of richly-annotated shapes represented by 3D CAD models of objects from 55 common categories. To generate the random room, we first need to randomly sample multiple objects from the the dataset. The number of objects we sample is a random integer from 12 to 18, which is similar to the average number of objects in ScanNetV2 scenes. Then for each sampled object, we perform the random room generation algorithm mentioned in Section~\ref{sec:construction}. The object-level contrastive learning loss is used to train the model in an unsupervised manner. 

For the downstream 3D object detection task, we use the backbone models proposed in~\cite{qi2019deep} and~\cite{zhang2020h3dnet}, which take as input 40,000 points. Following the network configurations in these two works, we use the 1024-point feature as the output of the backbone models and perform contrastive learning on this feature. During pre-training, we use the Adam optimizer~\cite{adam-kingma2014adam} with initial learning 0.001. We train the model for 300 epochs and the learning rate is multiplied by 0.1 at the 100-th and 200-th epcoh. The batch size is set to 16 such that roughly 200$\sim$300 unique objects are involved in the contrastive learning at every iteration.  

\subsection{3D Object Detection}

\begin{table*}[t!]
\caption{3D object detection results on ScanNetV2 validation set. Per-category results of average precision (AP) with IOU threshold 0.25 are reported.  We also show the mean of AP across all semantic classes with IoU threshold 0.25.} 
\newcolumntype{g}{>{\columncolor{Gray}}c}
\begin{adjustbox}{width=\textwidth}
 \begin{tabular}{r|c|*{18}{c}|g} 
 \toprule
 & Input & cab & bed & chair & sofa & tabl & door & wind & bkshf & pic & cntr & desk & curt & fridg & showr & toil & sink & bath & ofurn & mAP\\
 \midrule
3DSIS-5\cite{hou20193d} & Geo+RGB & 19.8 & 69.7 & 66.2 & 71.8 & 36.1 & 30.6 & 10.9 & 27.3 & 0.0 & 10.0 & 46.9 & 14.1 & 53.8 & 36.0 & 87.6 & 43.0 & 84.3 & 16.2 & 40.2 \\ 
3DSIS\cite{hou20193d} & Geo & 12.8 & 63.1 & 66.0 & 46.3 & 26.9 & 8.0 & 2.8 & 2.3 & 0.0 & 6.9 & 33.3 & 2.5 & 10.4 & 12.2 & 74.5 & 22.9 & 58.7 & 7.1 & 25.4  \\
 \midrule
Votenet\cite{qi2019deep} & Geo & 36.3 & 87.9 & 88.7 & 89.6 & 58.8 & 47.3 & 38.1 & 44.6 & 7.8 & 56.1 & 71.7 & 47.2 & 45.4 & 57.1 & 94.9 & 54.7 & 92.1 & 37.2 & 58.6 \\
Ours + VoteNet & Geo & 37.2 & 87.4 & 88.9 & 89.8 & 61.9 & 45.3 & 42.6 & 53.5 & 7.8 & 51.7 & 67.2 & 53.5 & 54.0 & 66.4 & 96.8 & 62.6 & 92.0 & 43.6 & 61.3 \\
\midrule
H3DNet\cite{zhang2020h3dnet} & Geo & 49.4 & 88.6 & 91.8 & {90.2} & 64.9 & {61.0} & 51.9 & {54.9} & {18.6} & {62.0} & 75.9 & {57.3} & 57.2 & 75.3 & 97.9 & 67.4 & {92.5} & 53.6 & 67.2 \\
Ours + H3DNet & Geo & {53.6} & {89.7} & {92.1} & 90.1 & {71.5} & 58.2 & {54.2} & 53.0 & 16.6 & 60.5 & {79.1} & 56.1 & {58.1} & {85.0} & {98.8} & {71.1} & 89.5 & {57.4} & \textbf{68.6} \\
\bottomrule
\end{tabular}
\label{Table:Quantitative:Result:ScanNetCat}
\end{adjustbox}
\end{table*}

\begin{table}
\caption{3D object detection results on ScanNetV2 validation set. We show mean of average precision (mAP) across all semantic classes with 3D IoU threshold 0.25 and 0.5. }
\newcolumntype{g}{>{\columncolor{Gray}}c}
\begin{adjustbox}{width=0.9\columnwidth, center}
\centering
 \begin{tabular}{r | c | g | g  } 
 \toprule
 & Input  &  mAP$_{25}$ & mAP$_{50}$  \\
 \midrule
 DSS\cite{song2016deep} & Geo $+$ RGB & 15.2 & 6.8 \\ 
 F-PointNet\cite{qi2018frustum} & Geo + RGB & 19.8 & 10.8 \\
 GSPN\cite{yi2019gspn} & Geo + RGB & 30.6 & 17.7\\
 3D-SIS \cite{hou20193d} & Geo + 5 views & 40.2 & 22.5 \\
 \midrule
 PointContrast~\cite{xie2020pointcontrast} & Geo only &  58.5 & 38.0  \\
  \midrule
 VoteNet \cite{qi2019deep} & Geo only & 58.6 & 33.5 \\
Ours + VoteNet & Geo only & 61.3 & 36.2 \\
  \midrule
   H3DNet~\cite{zhang2020h3dnet} & Geo only & 67.2 & 48.1 \\
 Ours + H3DNet & Geo only & 68.6 & 51.5 \\ 
 \bottomrule
\end{tabular}
\label{Table:scannet:map:0.5}
\end{adjustbox}
\end{table}

\noindent\textbf{Datasets.}
We conduct experiments on two widely-used 3D detection benchmarks, ScanNetV2~\cite{dai2017scannet} and SUN-RGBD~\cite{song2015sun}. ScanNetV2 is a richly annotated dataset of 3D reconstructed meshes of indoor scenes. It contains 1,513 scanned and reconstructed real scenes, which consists of 18 different categories of objects of various size and shape. Currently, it is the largest one that was created with a light-weight RGB-D scanning procedure. Yet, it is still much smaller in scale when compared to datasets in 2D vision. We split the the whole dataset into two subsets with 1,201 and 312 scenes for training and testing following~\cite{qi2019deep,dai2017scannet}. SUN RGB-D is a single-view RGB-D dataset for 3D scene understanding. It contains of 10,335 indoor RGB and depth images with object bounding boxes and per-point semantic labels with 10 different categories of objects. We also strictly follow the splits described in~\cite{qi2019deep,dai2017scannet}, with 5,285 samples as training data and 5,050 as testing data. 

\vspace{5pt} \noindent \textbf{Detection Models.}
We compare our method with two recently proposed state-of-the-art approaches: One is VoteNet~\cite{qi2019deep}, which is a geometric-only detector that combines deep point set networks and a voting procedure; the other is H3DNet, which predicts a hybrid set of geometric primitives. Both of them take colorless 3D point clouds as input.
We also include GSPN~\cite{yi2019gspn}, 3D-SIS~\cite{hou20193d}, DSS~\cite{song2016deep}, F-PointNet~\cite{qi2018frustum}, 2D-Driven~\cite{lahoud20172d}, and Cloud of gradient (COG)~\cite{ren2016three}, which use other types of information for object detection, into the comparison.

\vspace{5pt} \noindent \textbf{Implementation Details. }
We show the effectiveness of our method by the improvement upon VoteNet and H3DNet. We load the pre-trained part into the model at the beginning of the training, and follow their training setting. Specifically, we train the model for 360 iterations in total. The initial learning is 1e-2 and 1e-3 for ScanNetV2 and SUN-RGBD respectively. We evaluate the performance by mAP with 3D IoU threshold as 0.25 and 0.5. Please refer the original paper for more details with regard to the experimental settings.

\vspace{5pt} \noindent \textbf{ScanNetV2. }
We first report the results of mAP@0.25 as well as AP@0.25 for all semantic classes in Table~\ref{Table:Quantitative:Result:ScanNetCat}. With the pre-training, we improve the mAP by 2.6 point and 1.4 points for VoteNet and H3DNet respectively. These results indicate that our pre-training can truly improve the fine-tuning on high-level detection tasks. Moreover, for 11 out of 18 categories, improvement of the average precision can be observed. This indicates the pre-training can boost the detection of most common categories.

We further report the results of mAP@0.5, which is a more difficult metric, and add the comparison with other 3D object detection approaches that utilize the color information in Table~\ref{Table:scannet:map:0.5}. For both mAP@0.25 and mAP@0.5 metric, our method achieves the state-of-the-art. In particular, for mAP@0.5, the improvement is even larger than mAP@0.25, that is, we improve by 2.7 points and 3.4 points upon VoteNet and H3DNet respectively. This indicates we can obtain more accurate bounding box prediction with the help of proposed pre-training strategy.

\vspace{5pt} \noindent \textbf{SUN RGB-D. }
We also conduct the experiments on SUN RGB-D. We report the results in Table~\ref{Table:Quantitative:Result:SUN}. With pre-training, we again achieve the state-of-the-art. For mAP@0.25, we improve 1.5 points for both VoteNet and H3DNet. For mAP@0.5, we improve 2.5 points and 4.1 points for VoteNet and H3DNet. This result once again illustrates our method can predict more accurate bounding box. As for the average precision of each class, improvement can be observed for 7 out of 10 categories.

\begin{table*}
\caption{3D object detection results on SUN RGB-D val dataset. We report per-category results of average precision (AP) with 3D IoU threshold 0.25, and mean of AP across all semantic classes with 3D IoU threshold 0.25 and 0.5. For fair comparison, with previous methods, the evaluation is on the SUN RGB-D V1 data.}
\label{Table:Quantitative:Result:SUN} \centering
\newcolumntype{g}{>{\columncolor{Gray}}c}
\begin{adjustbox}{width=0.95\textwidth}
 \begin{tabular}{r | c | *{10}{c} | g | g } 
\toprule
 & Input & bathtub & bed & bkshf & chair & desk & drser & nigtstd &sofa &table & toilet & mAP$_{25}$ & mAP$_{50}$  \\
\midrule
 DSS\cite{song2016deep} & Geo + RGB & 44.2 & 78.8 & 11.9 & 61.2 & 20.5 & 6.4& 15.4& 53.5& 50.3& 78.9 & 42.1 & - \\ 
 COG\cite{ren2016three} & Geo + RGB & 58.3 & 63.7 & 31.8 & 62.2 & 45.2 & 15.5 & 27.4& 51.0 & 51.3 & 70.1 & 47.6 & -\\ 
 2D-driven\cite{lahoud20172d} & Geo + RGB & 43.5 & 64.5 & 31.4 & 48.3 & 27.9 & 25.9 & 41.9 & 50.4 & 37.0 & 80.4 & 45.1 & - \\
 F-PointNet\cite{qi2018frustum} & Geo + RGB & 43.3 & 81.1 & 33.3 & 64.2 & 24.7 & 32.0 & 58.1 & 61.1 & 51.1 & 90.9 & 54.0 & - \\
\midrule
PointContrast~\cite{xie2020pointcontrast} & Geo & - & - & - & - & - & - & - & - & - & - & 57.5 & 34.8 \\
\midrule
 VoteNet \cite{qi2019deep} & Geo & 74.7 & 83.0 & 28.8 & 75.3 & 22.0 & 29.8 & 62.2 & 64.0 & 47.3 & 90.1 & 57.7 & 32.9 \\
 Ours + VoteNet  & Geo & 76.2 & 83.5 & 29.2 & 76.7 & 25.1 & 33.2 & 64.2 & 63.8 & 49.0 & 91.2 & 59.2 & 35.4 \\
\midrule
 H3DNet~\cite{zhang12356deep} & Geo & {73.8} & 85.6 & 31.0 & 76.7 & {29.6} & 33.4 & 65.5 & 66.5 & {50.8} & 88.2 & 60.1 & 39.0 \\
 Ours + H3DNet & Geo & 71.2 & {86.4} & {38.7} & {77.8} & 28.0 & {36.5} & {68.3} & {67.7} & 50.3 & {91.0} & \textbf{61.6} & \textbf{43.1} \\
\bottomrule
\end{tabular}
\end{adjustbox}
\end{table*}

\vspace{5pt} \noindent \textbf{Less Training Data. }
To show our method can truly learn a better initialization through pre-training, we further conduct the empirical studies with much less training data. We report the results in Table~\ref{Table:few-shot}. We use 5$\%$, 10$\%$, 25$\%$ and 50$\%$ of the training data from ScanNetV2 dataset. As can be seen from the Table~\ref{Table:few-shot}, the improvement under this few-shot setting is still obvious, especially for mAP@0.25. The improvement on mAP@0.25 is even growing larger when less data is used. Notably, the improvement of mAP@0.25 is larger than 5 points when we use less than 10$\%$ training data. On the other hand, the improvement on mAP@0.5 is almost unchanged compared to mAP@0.25. This indicates our pre-training method can help the model of downstream high-level tasks to achieve a better coarse understanding of the scene when less data is available. But to gain more accurate understanding, we still need supervised learning with annotated data.

\begin{table*}
\caption{Effects of the training data size. We show the mean of AP across all semantic classes with 3D IoU threshold 0.25 and 0.5 when training on ScanNetV2 with less data. We report the results of using 5$\%$, 10$\%$, 25$\%$ and 50$\%$ data.}
\centering
\begin{adjustbox}{width=0.9\textwidth}
 \begin{tabular}{r | cc | cc | cc | cc | cc  } 
\toprule
 &  \multicolumn{2}{c|}{100\%} & \multicolumn{2}{c|}{50\%} & \multicolumn{2}{c|}{25\%} & \multicolumn{2}{c|}{10\%} & \multicolumn{2}{c}{5\%}\\
 & mAP$_{25}$ & mAP$_{50}$ &  mAP$_{25}$ & mAP$_{50}$ &  mAP$_{25}$ & mAP$_{50}$ &  mAP$_{25}$ & mAP$_{50}$ &  mAP$_{25}$ & mAP$_{50}$  \\
\midrule
VoteNet \cite{qi2019deep} & 58.6 & 33.5 & 47.0 & 25.3 & 35.5 & 20.0 & 25.1 & 14.3 & 12.6 & 3.2\\
\rowcolor{Gray} Ours + VoteNet &  61.3 & 36.2 & 53.0 & 30.2 & 38.2 & 23.2 & 28.9 & 17.2 & 19.1 & 10.1\\
\midrule
H3DNet~\cite{zhang12356deep}& 67.2 & 48.1 & 61.5 & 40.6 & 51.6 & 30.9 & 37.0 & 20.7 & 26.6 & 11.3  \\
\rowcolor{Gray} Ours + H3DNet & 68.6 & 51.5 & 63.2 & 43.6 & 54.4 & 33.5 & 42.2 & 23.4 & 32.0 & 13.9 \\
\bottomrule
\end{tabular}
\label{Table:few-shot}
\end{adjustbox}
\end{table*}

\vspace{5pt} \noindent \textbf{Ablation Study.}
In Table~\ref{Tab:ablation}, we conduct three groups ablation studies. All these ablation studies are conducted on ScanNetV2 dataset with VoteNet as the backbone. We use mAP@0.25 as the evaluation metric.

We first study the choice of datasets where the pre-training is performed. From Table~\ref{Tab:ablation:a},  we observe that pre-training on either ShapeNet or ScanNetV2 can both improve the performance. Yet, thanks to the larger scale of ShapeNet, i.e. more samples from more diverse categories, pre-training on it can achieve better results compared to ScanNetV2. Furthermore, we exhibit the possibility to combine both datasets to help the pre-training. Having the objects from both datasets, we can achieve even better fine-tuning result compared to one single dataset is used.

We then study the effect of loss function used for pre-training in Table~\ref{Tab:ablation:b}. Compared to the point-level contrastive loss used by PointContrast, we can achieve even better pre-training results with the instance-level contrastive loss. This indicates the object-level contrastive learning can  better help the downstream localization tasks by incorporating more instance-level knowledge. Considering that the label of objects in ShapeNet is easy to access, we also add an additional segmentation loss by assigning all the points of an object with the corresponding object label. This can bring some marginal improvement with additional supervision signal being used. This illustrates the fact that our complete unsupervised pre-training strategy can achieve comparable performance with the supervised pre-training on synthetic dataset. 

We finally show the necessity of some strategies used in scene generation. In Table~\ref{Tab:ablation:c}, we verify the necessity of gravity principle and the need of floor and wall in a scene. Without these components, we can still improve upon the baseline, but the larger domain shift between real scene and generated scene may hamper the pre-training from obtaining better model for fine-tuning on the real dataset of downstream tasks.

\begin{table*}
\caption{Ablation analysis on the proposed RandomRooms method. We investigate the effects of pre-training datasets, learning losses and random room generation methods. We report the mAP$_{25}$ results of VoteNet on ScanNetV2. }
\centering
\subfloat[Ablation studies on pre-training datasets.]
{\makebox[0.3\linewidth][c]{
\tablestyle{12pt}{1.2}
\begin{tabular}{c|c}
\textbf{Pre-training dataset} & \textbf{mAP} \\
\shline
 baseline & 58.6 \\
\hline
ScanNetV2 & 60.2 \\
\rowcolor{Gray} ShapeNet & 61.3 \\ 
ShapeNet + ScanNetV2 & 61.5 \\
\end{tabular}
\label{Tab:ablation:a}
} 
}
\hfill
\subfloat[Ablation studies on pre-training losses.\label{tb:method}]
{\makebox[0.3\linewidth][c]{
\tablestyle{12pt}{1.2}
\begin{tabular}{c|c}
\textbf{Pre-training loss} & \textbf{mAP} \\
\shline
 baseline & 58.6 \\
\hline
point-level contrastive & 59.2 \\
\rowcolor{Gray} instance-level contrastive  & 61.3 \\ 
instance-level contrastive + seg. & 61.5 \\
\end{tabular}
\label{Tab:ablation:b}
}
}
\hfill
\subfloat[Ablation studies on room generation.
\label{tab:analysis:pixel}]
{\makebox[0.3\linewidth][c]{
\tablestyle{12pt}{1.2}
\begin{tabular}{c|c}
\textbf{Generation method} & \textbf{mAP} \\
\shline
 baseline & 58.6 \\
\hline
\rowcolor{Gray} RandomRooms  & 61.3 \\ 
w/o gravity & 60.5 \\
w/o floor/wall & 60.7 \\
\end{tabular}
\label{Tab:ablation:c}
} 
}
\vspace{-5pt}
\label{Tab:ablation}
\end{table*}

\begin{figure}
  \centering
  \includegraphics[width=\linewidth]{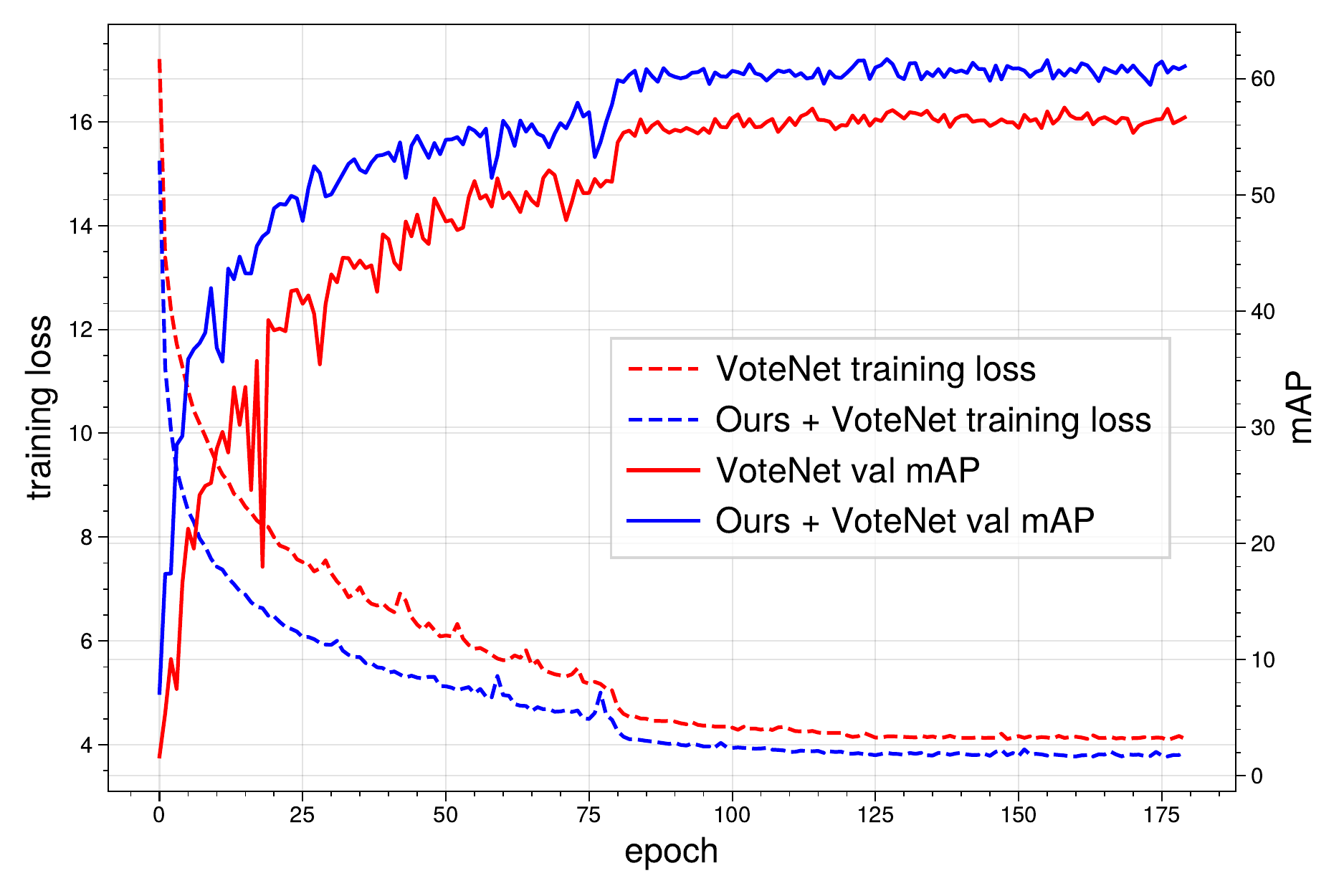}
  \caption{Training from scratch vs. fine-tuning with RandomRooms pre-trained weights. We report the 3D detection training loss and the validation mAP@0.25 of VoteNet on ScanNetV2. }
  \label{fig:curve}
  \vspace{-10pt}
\end{figure}

\begin{table}
\caption{We compare our method with PointContrast on ScanNetV2 and SUN R-GBD using PointNet++ as backbone. We show mean of average precision (mAP) across all semantic classes with 3D IoU threshold 0.25. }
\newcolumntype{g}{>{\columncolor{Gray}}c}
\begin{adjustbox}{width=\columnwidth, center}
\centering
 \begin{tabular}{r | g |  g } 
 \toprule
  &  ScanNetV2 & SUN RGB-D  \\
 \midrule
 Sparse Res-UNet w/o pre-training & 56.7 & 55.6\\
 Sparse Res-UNet w/ PointContrast &  58.5 & 57.5 \\
  \midrule
 PointNet++ w/o pre-training & 58.6 & 57.7\\
 PointNet++ w/ PointContrast &  58.5 & 57.9 \\ 
 \midrule
 PointNet++ w/ RandomRooms & \textbf{61.3} & \textbf{59.2} \\
 \bottomrule
\end{tabular}
\label{Table:scannet:pretrain:0.25}
\end{adjustbox}
\end{table}

\vspace{5pt} \noindent \textbf{Comparison with PointContrast. } To show our pre-training method is more suitable for the 3D object detection task, we compare with another pre-training method, namely PointContrast, on ScanNetV2 and SUN RGB-D using VoteNet~\cite{qi2019deep} as the detection model, and we use mAP@0.25 as the evaluation metric. The results are reported in Table~\ref{Table:scannet:pretrain:0.25}

We find that using Sparse Res-UNet instead of PointNet++ as the backbone model leads to worse detection performance when training from scratch. However, the improvement brought by PointContrast to the detectors based on PointNet++ is quite marginal, and the final performance is on par with the detectors using Sparse Res-UNet as the backbone. On the contrary, considering there is no need to keep the point correspondence, our RandomRooms method can learn a much better initialization for the PointNet++ style model, which is stronger backbone for current state-of-the-art 3D object detectors. This demonstrates our method is superior on the object detection task compared to PointContrast.

\vspace{5pt} \noindent \textbf{Learning Curve. } We show the learning curve of our method as well as the baseline VoteNet in Figure~\ref{fig:curve}. We observe that our pre-trainig weights significantly help improve the learning speed and stabilize the training process.  The model with pre-training weights can achieve lower training loss and better validation mAP, which clearly demonstrates the effectiveness of the proposed method.

\begin{figure}
  \centering
  \includegraphics[width=\linewidth]{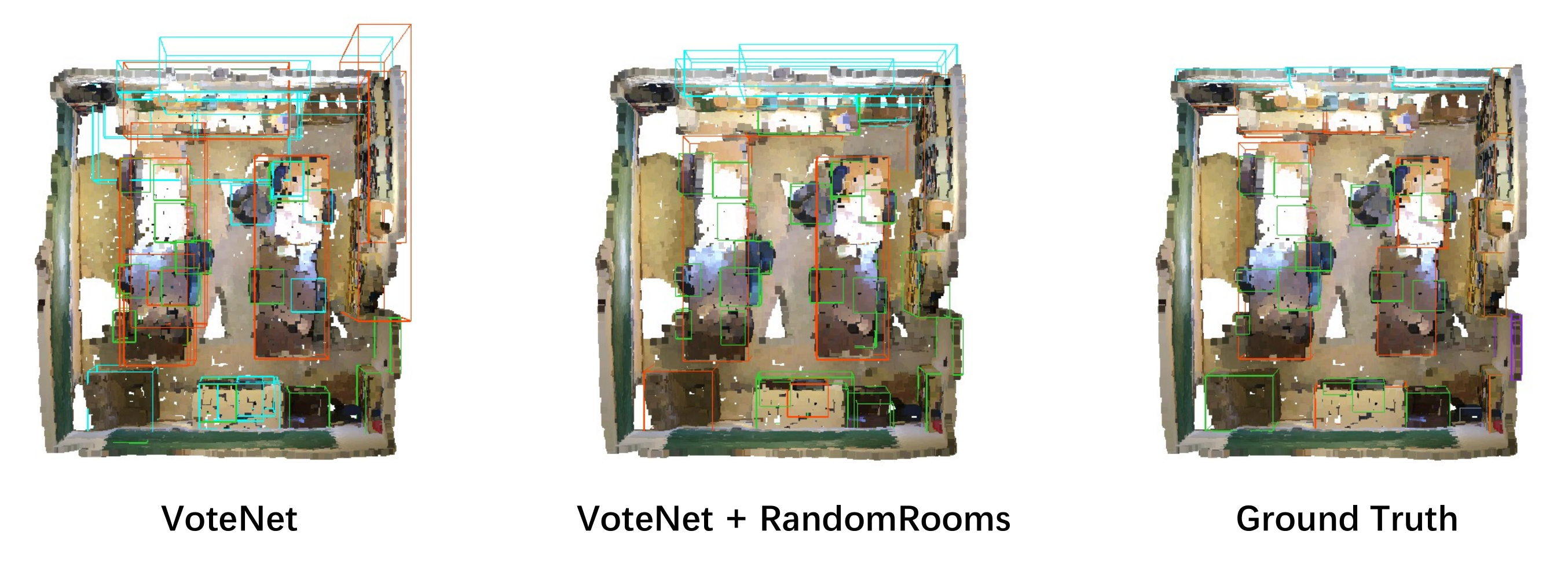}
  \caption{Visual Results on ScanNetV2. We compare the qualitative detection results with the baseline VoteNet method. The pre-trained model can produce more accurate detection results with less false positives, and is closer to the ground-truth bounding boxes. }
  \label{fig:vis_det}
  \vspace{-10pt}
\end{figure}

\vspace{5pt} \noindent \textbf{Visualization. } We visualize the detection results of the baseline VoteNet that is trained from scarch and the pretrained model using our method on ScanNet. The results are shown in Figure~\ref{fig:vis_det}.  We see the pre-trained model can produce more accurate detection results with less false positives, and is closer to the ground-truth bounding boxes. The visual results further confirm the effectiveness of the proposed method. 

\vspace{5pt} \noindent \textbf{Discussions.}
Though we follow many heuristic rules when generating the \emph{random rooms}, there still exist domain gap between the real scene and generated one. The extensive experimental results shed light on an interesting fact, that is, in 3D representation learning the layout of objects may not be that important for recognition as in 2D vision. We only need to ensure the set of objects can spread out in the space, while the interaction among objects does not matter that much as 2D vision where hidden interactions may play as an important cue for many high-level scene understanding tasks like detection. This may be due to the ovelap is not that severe in complex 3D scenes. We think this may open a path for future research on 3D learning.

\section{Conclusion}
In this paper, we have proposed a new pipeline, namely RandomRoom, for 3D pre-training that can make use of the synthetic CAD model dataset to help the learning on real dataset on high-level 3D object detection task. Unlike previous works performing contrastive learning at the level of points, we perform  contrastive learning at the object level by composing two different scenes with same set of objects that are randomly sampled from the CAD model dataset.
Empirically, we show consistent improvements in downstream 3D detection tasks on several base models, especially when less training data are used. Benefiting from the rich semantic knowledge and diverse objects from synthetic data, our method establishes the new state-of-the-art on widely-used 3D detection benchmarks ScanNetV2 and SUN RGB-D. We except this work can open a new path for future research on how to exploit easily accessible synthetic 
objects for more complex tasks for 3D scene understanding. 

\subsection*{Acknowledgements}
This work was supported in part by the National Key Research and Development Program of China under Grant 2017YFA0700802, in part by the National Natural Science Foundation of China under Grant 61822603, Grant U1813218, and Grant U1713214, in part by a grant from the Beijing Academy of Artificial Intelligence (BAAI), and in part by a grant from the Institute for Guo Qiang, Tsinghua University.

\section*{A. Details about Random Room Generation}

To more clearly show the generation process of random rooms, we provide pseudo-code and explanatory
comments of our room generation method in Algorithm~\ref{alg:supp}.

\begin{algorithm*}[p]
\caption{Pseudo-code of Random Room Generation}
\label{alg:supp}

\begin{lstlisting}[language=Python]
# objects: list of object point clouds

# object level data augmentation
objcts = object_augmentation(objcts)

# sort objects by their areas
objects, object_ind, obj_area = sort_object(objects)

# set the overall area of the rectangular room
overall_area = sum(obj_area) * 2 * 10000 * (random.random() * 0.4 + 0.6)
a_value = np.sqrt(overall_area)
a = random.randint(int(a_value*0.75), int(a_value*1.25))
b = int(overall_area) // a
a_m = float(a) / 100.
b_m = float(b) / 100.
room_state = np.zeros((a, b), dtype=np.float)

final_layout = []
instance_label = []

# place object to the room
for i in range(len(objcts)):
    obj = objects[i]
    x, y, z = get_object_size(obj)

    for _ in range(max_iter):
        # generate the position from beta distribution
        pos_x = np.random.beta(0.5, 0.5) * (a_m - x)
        pos_y = np.random.beta(0.5, 0.5) * (b_m - y)
        state_part = room_state[int(pos_x*100):int((pos_x+x)*100), int(pos_y*100): int((pos_y+y)*100)]
        max_height = state_part.max()
        if (max_height + z < 2.0 and max_height < 0.5) or max_height < 1e-3:
            break
    
    room_state[int(pos_x * 100):int((pos_x + x) * 100), int(pos_y * 100): int((pos_y + y) * 100)] += z
    obj[:, 0] += pos_x
    obj[:, 1] += pos_y
    obj[:, 2] += max_height
    final_layout.append(obj)
    instance_label.append(np.ones((obj.shape[0],), dtype=int) * (object_ind[i] + 1))

# add floor and walls 
final_layout, instance_label = add_floor_wall(final_layout, instance_label)

# form the final scene point cloud
final_layout = np.concatenate(final_layout, axis=0)
instance_label = np.concatenate(instance_label, axis=0)

# normalize coordinates
final_layout[:, 0:2] = final_layout[:, 0:2] - final_layout[:, 0:2].mean(axis=0, keepdims=True)
\end{lstlisting}

\end{algorithm*}

\section*{B. Visualization of Random Rooms}

We show more examples of the generated random room pairs in Figure~\ref{fig:example}. 

\begin{figure*}[p]
  \centering
  \includegraphics[width=0.95\linewidth]{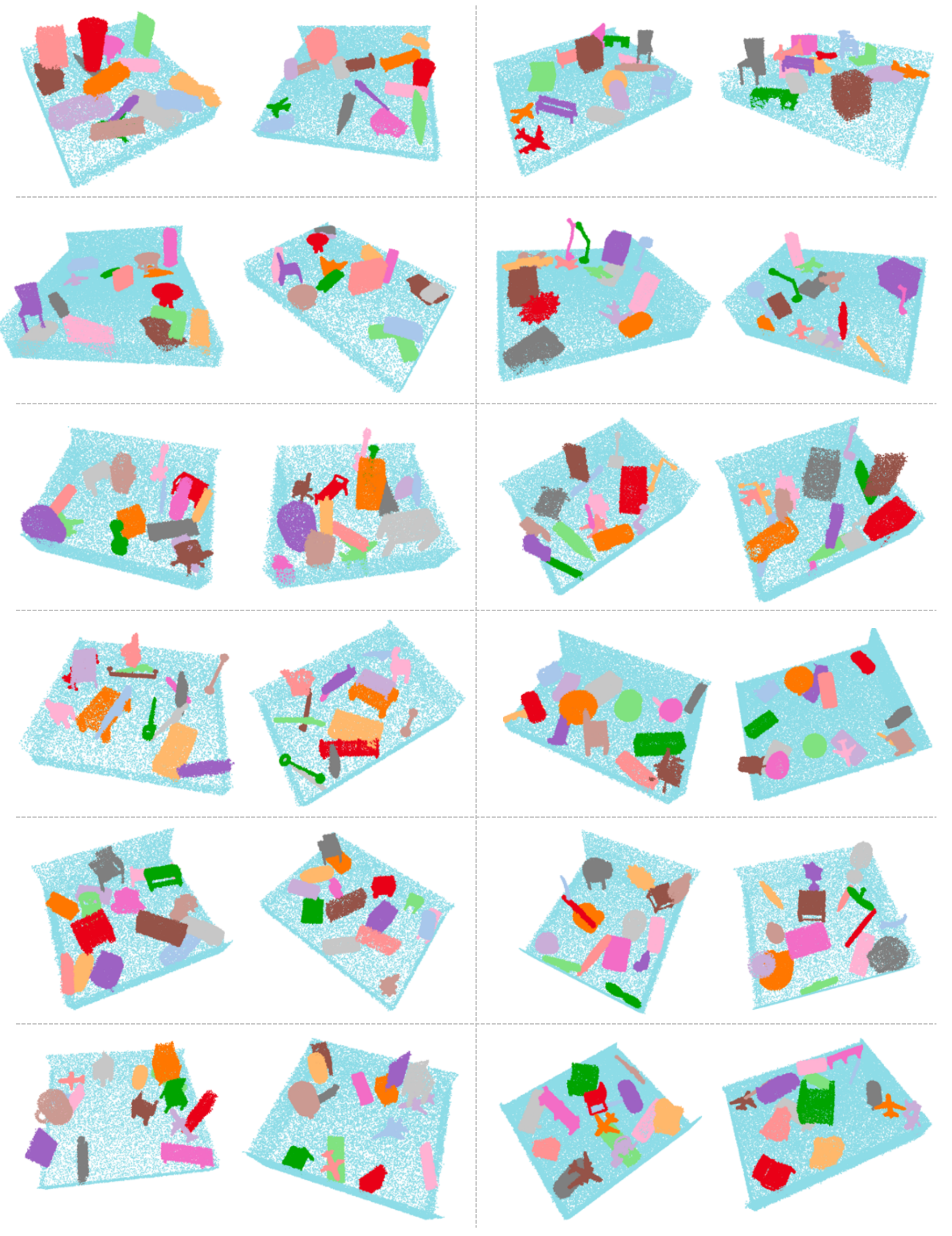}
  \caption{Visualization of the pairs of \textit{Random Rooms}. }
  \label{fig:example}
\end{figure*}

{\small
\bibliographystyle{ieee_fullname}
\bibliography{reference}
}

\end{document}